# Deep learning based automatic segmentation of lumbosacral nerves on non-contrast CT for radiographic evaluation: a pilot study

Guoxin Fan[1,2,3,#], Huaqing Liu[2,#], Zhenhua Wu[4], Yufeng Li[5], Chaobo Feng[1,2], Dongdong Wang[1,2], Jie Luo[3,6], Xiaofei Guan[7], William M. Wells III[3], Shisheng He[1,2,*].

[1] Shanghai Tenth People's Hospital, Tongji University School of Medicine, Shanghai, China
[2] Spinal Pain Research Institute of Tongji University, Shanghai, China
[3] Surgical Planning Lab, Radiology Department, Brigham and Women's Hospital, Harvard Medical School, MA, USA
[4] School of Data and Computer Science, Sun Yat-sen University, Guangzhou, China
[5] Shanghai Jiao Tong University School of Medicine, Shanghai, China
[6] Graduate School of Frontier Sciences, The University of Tokyo, Tokyo, Japan
[7] Burke Neurological Institute, White Plains, NY, USA.

# these two authors equally contribute to the study.
* corresponding authors: Shisheng He (tjhss7418@tongji.edu.cn)

[Abstract]
Background and objective: Combined evaluation of lumbosacral structures (e.g. nerves, bone) on multimodal radiographic images is routinely conducted prior to spinal surgery and interventional procedures. Generally, magnetic resonance imaging is conducted to differentiate nerves, while computed tomography (CT) is used to observe bony structures. The aim of this study is to investigate the feasibility of automatically segmenting lumbosacral structures (e.g. nerves & bone) on non-contrast CT with deep learning.
Methods: a total of 50 cases with spinal CT were manually labeled for lumbosacral nerves and bone with Slicer 4.8. The ratio of training: validation: testing is 32:8:10. A 3D-Unet is adopted to build the model SPINECT for automatically segmenting lumbosacral structures. Pixel accuracy, IoU, and Dice score are used to assess the segmentation performance of lumbosacral structures.
Results: the testing results reveals successful segmentation of lumbosacral bone and nerve on CT. The average pixel accuracy is 0.940 for bone and 0.918 for nerve. The average IoU is 0.897 for bone and 0.827 for nerve. The dice score is 0.945 for bone and 0.905 for nerve.
Conclusions: this pilot study indicated that automatic segmenting lumbosacral structures (nerves and bone) on non-contrast CT is feasible and may have utility for planning and navigating spinal interventions and surgery.

## Introduction
Low back pain is a common ailment in clinics, which is a global leading cause to disability and adds a great burden to healthcare expense [1][2]. It is difficult to reveal the etiology of a specific low back pain, but it is usually caused by the inflammation of nerves from a mechanic compression or chemical irritation [3]. In practice, spinal intervention and spinal surgery are often recommended for the management of low back pain, and all these procedures heavily rely on medical image guidance. Spinal intervention and spinal surgery at L5/S1 level is problematic due to anatomic obstacles such as high iliac crest, enlarged transverse process, and limited foraminal area [4][5]. Thus, pre-procedure radiographic evaluation is a key to success, and image-based navigation or robotic procedures may also improve clinical efficiency.

Generally, spinal intervention and surgery are conducted under the guidance of X-ray fluoroscopy, ultrasound, magnetic resonance imaging (MRI) and computed tomography (CT) [6]. Although fluoroscopy is the gold standard for image guidance, it is incapable of providing information about soft tissues. MRI and ultrasound are newly developed image modalities for spinal intervention, but the accuracy and safety need further validation [6]–[9]. CT or CT-fluoroscopy is an emerging tool for guiding spinal intervention and surgery, because they can quickly and safely localize needle or other instruments and minimize risk of nerve injury [10]–[12]. As a result, we select thin-layer CT of the lumbosacral segment to construct a 3D model for radiographic evaluation. However, precise segmentation is needed for 3D reconstruction, and there is no studies achieving automatic segmentation of lumbosacral nerves on CT to the best of our knowledge.

Recently, deep learning has gained substantial attention in the field of radiology [13]–[16]. Deep learning algorithms are able to learn from large amounts of data using neural networks, frequently convolutional neural networks (CNNs) [17]. Although CNNs were proposed decades ago, it is the recent 6 years that deep learning has achieved great success due to massive available data, increased processing power, and rapid development of algorithms [17]. The U-Net is a kind of CNN that was developed for biomedical image segmentation by Ronneberger et al. in 2015, and it has many applications in segmentation of two-dimensional images [18]. In 2016, the same group developed 3D U-net for volumetric segmentation as an extension architecture of U-net [19]. Many studies have validated the outstanding performance of 3D U-net in segmentation of volumetric medical image [20]–[22].

The aim of the study is to investigate the feasibility of automatically segmenting lumbosacral structures (nerves and bone) on non-contrast CT with a 3D U-net.

**Methods**

This retrospective study was approved by the local institutional ethical committee prior to data extraction. All the CT data was obtained from Shanghai Tenth People's Hospital, and all algorithm were developed and tested using Keras (V2.1.1 with Tensorflow back-end) in a personal computer (GPU: an Nvidia GeForce GTX 960 with 4 GB of memory; an 3.5GHz Intel(R) Core(TM) i7-4790 CPU with 8 GB of memory ).

**A. Manual annotations**

All included data was manually segmented with Slicer 4.8 [23]. Lumbosacral nerves and bones were meticulously segmented and labeled (**Figure 1**). These manual annotations were regarded as the ground truth.

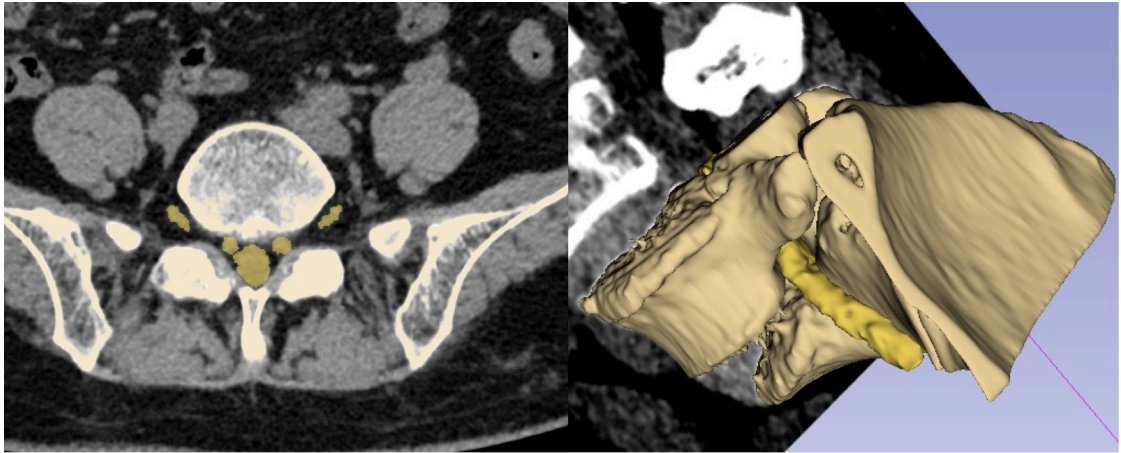

Figure 1. Manual segmentation and 3D reconstruction of nerves and bone on Slicer (left: manual labels; right: 3D reconstruction with coronal image).

**B. Data preprocessing**

All thin-layer CT were preprocessed using the following steps: resampling, cropping, and intensity normalization.

1) Resampling: we standardized the voxel size of original CT and manually labeled 3D masked images into 1mm×1mm×1mm nearest neighbor interpolation.
2) Cropping: we cropped the resampled CT and manually-labeled 3D masked images into 32×64×64 patches, which were used as input data to train the 3D-Unet.
3) Intensity normalization: For more accurate semantic segmentation, the average brightness and contrast fluctuations of CT images of different samples should have a degree of consistency. For this reason, the CT images in the dataset are standardized so that the pixel values of the CT images have zero mean and unit variance. The scale and bias were obtained by statistic computing from training dataset, and they were used for whitening during all phases including training, validation and testing.
4) Data augmentation: we conducted data augmentation with the following methods: 1. Adding a small amount of white noise to patch, which will be input to the neural network, 2. Performing vertical flipping and horizontal flipping at a probability of 0.5; 3. Voxel size is randomly disturbed within the range of ±0.2 mm to introduce a degree of size variation.

**C. Network architecture**

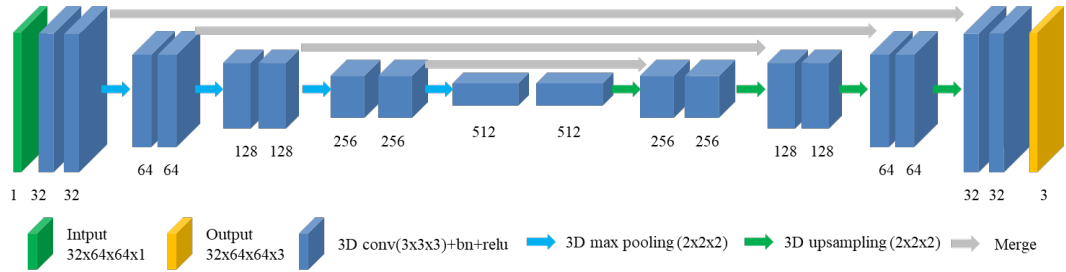

Figure 2. Schematic of the network architecture.

The 3D-Unet is adopted for multi-class segmentation of lumbosacral structures (Figure 2). The adopted network consists of two parts including the encoder part and the decoder part. The encoder part performs data analysis and feature representation learning from the input data and the decoder part generates segmentation results.

There are also 4 shortcut connections between layers of equal resolution in the encoder and decoder paths. The last layer of the model is 1×1×1 convolutional layer followed by a softmax layer, and the number of output channels is 3. The input of the model is 32×64×64 voxel tile of CT. The output is the corresponding probability mask, and its shape is 32×64×64×3. The whole architecture has 22581411 parameters. The developed model will be called SPINECT, as it aims to automatically segment multiple structures solely on spinal CT.

**D. Training**

The ratio of training, validation and testing is 32:8:10. Due to voxel disparity among background, bone and nerves in CT images, the segmentation accuracy of nerves with smaller voxel proportion will be very low if the conventional loss function (e.g. cross entropy loss function) is adopted, Thus, we used the following weighted softmax cross entropy loss function for training:

$$L = -\sum_{x \in \Omega} w(x) \log \left( \frac{\exp(a_l(x))}{\sum_{k=0}^{N} \exp(a_k(x))} \right)$$

In the above function, $x \in \Omega$ is voxel point; $a_k(x)$ is the activation value at the last layer before softmax of voxel point $x$ in channel k; $a_l(x)$ is the activation value at the last layer before softmax of voxel point $x$ in the groundtruth channel; $w(x)$ is the cross entropy weight of voxel x. According to the manually labeled 3D mask, different categories of voxels x in the loss function are given different weights to balance the category frequencies. It was found that the greater the difference in cross entropy weight between different categories of voxels, the more unstable the training process and the slower the convergence speed. After balance is reached, the IoUs of the categories with greater weights such as nerves will be somewhat low. Conversely, if the difference in cross-entropy weights of different categories of voxels is small, the convergence rate during training is faster, but the category with fewer voxels (such as nerves) will be easily covered by other categories and may not even be correctly identified. After reaching equilibrium, the IoUs with fewer voxels are also lower. To this end, the following training strategy was designed. In the early stage of training (before epoch 40), the weight ratio of the three types of voxels (background:bone:nerve) is 1:1:20. After it becomes stable (the last 60 epoch), we reduced the weight difference to 1:1:2 to continue training to reach a new equilibrium state.

During the training phase, a number of mini-batch CT and manual labeling masks are randomly selected from the training dataset, and the image data is subjected to the standardization processing as described in B1) to B3) and the augmentation operation described in B4). As a result, the training data input to each training iteration is different, which improves the generalization ability of the model. The convolutional layer parameters of 3D-Unet are initialized by the He method [24]. The size of the patch (depth×height×width) input to the 3D-Unet during training is 32×64×64 (unit: voxel) and the mini-batch is 4, which is optimized by stochastic gradient descent algorithm and the learning rate is 5e-4. The trained network was called SPINECT, as it is trained from spinal CT. Pixel accuracy, IoU, and Dice score are used to assess the segmentation performance of lumbosacral structures.

### E. 5-fold Validation

During model training, one validation was performed every 100 training iterations. Specifically, 6 cases were randomly selected from the validation dataset for standard processing. The standard processing including several sequential steps: 1) set the voxel size of the three dimensions to 1 by the nearest interpolation method; 2) standardizing. Then, we add random noise plus random horizontal and vertical flipping to augment the selected cases. A sliding window of size 32×64×64 is used to traverse the case with stride=(20×40×40) to obtain the patch $x_i$. The patch $x_i$ will be input to the current model $M$, and then the model will generate the corresponding probability mask $y_i = M(x_i)$. Finally, the automatic segmented mask $S$ will be obtained with a algorithm combining $y_i$ based on the location of $x_i$ on $Y$. The Dice Score of each voxel class is obtained with comparison of automatic segmented mask and manually labeled mask. If the average of the Dice Score is greater than the prior best Dice Score by the current iteration, the current model parameters will be saved.

The combined algorithm is as follows. Based on the location $L_i$ of patch $x_i$ in the CT, the corresponding probability mask $y_i$ is accumulated to the appropriate location of $Y$ as $Y(L_i)$. For any voxel $v(d,h,w) \in X$, the response of different channels of the last dimension in $Y(d,h,w,:)$ will be compared. If the value of channel k is the largest, then the voxel belongs to category k. The combined algorithm is as follows in **Table 1**.

Table 1. Overview of the Combined algorithm.

| **Algorithm 1**: Overview of the Combined algorithm |
|---|
| **Require**: $X$: CT volume, shape = D×H×W |
| **Require**: $x_i = X(L_i), (i = 1,...,k)$: CT voxel patch |
| **Require**: $y_i = M(x_i)$: $y_i$ is the output of the last layer (softmax activation function) of the model $M$, $y_i$ has one more dimension than $x_i$, and this dimension has 3 channels. Each channel refers to the probability of the corresponding voxel belongs to background or bone or nerve, respectively. |
| 1: **Initiliaze**: $Y \leftarrow 0$ |
| 2: **for** $x_i \in X, (i = 1,...,k)$ **do** |
| 3:     $Y(L_i,:) += y_i$ |
| 4: **end for** |
| 5: $S \leftarrow \arg\max(Y, axis = -1)$ (find the channel with the largest value in the last dimension) |
| 6: **return** $S$ (the automatic mask) |

### F. Testing:

During testing, all cases in the test dataset were selected and underwent standard processing, and then a sliding window of size 32×64×64 is used to traverse the case with stride=(20×40×40) to obtain the patch $x_i$. The patch $x_i$ was input to the trained model $M$, and then the model generated the corresponding probability mask $y_i$.

Finally, the automatic segmented mask $S$ was obtained with the above combined algorithm combining $y_i$ based on the location of $x_i$ on $Y$.

**Results**

Testing results reveal that SPINECT can achieve successful segmentation of multiple structures (bone and nerve) on CT (Figure 3-5). Quantitative segmentation accuracy is shown in Table 2, Table 3 and Table 4. The average pixel accuracy is 0.940 for bone and 0.918 for nerve. The average IoU is 0.897 for bone and 0.827 for nerve. The dice score is 0.945 for bone and 0.905 for nerve. In each validation fold, it took about 4 hours and 35 minutes to finish the training of segmentation net. After training, it took on average about 3.1 seconds to finish the segmentation of one test CT data.

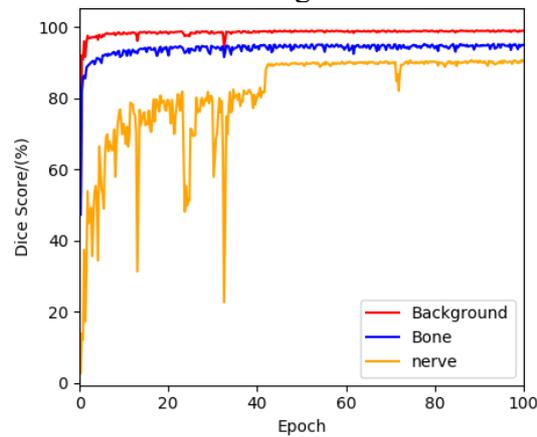

Figure 3. The convergence curve of Dice Score.

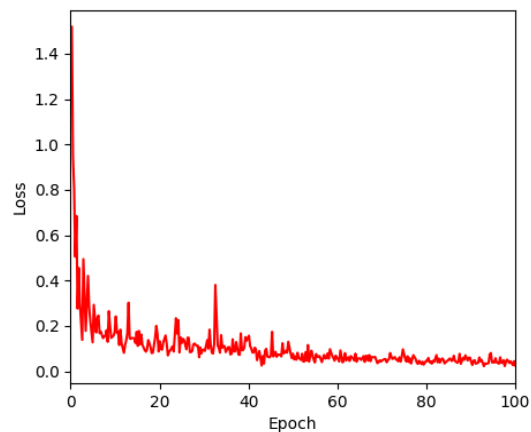

Figure 4. The value of loss function.

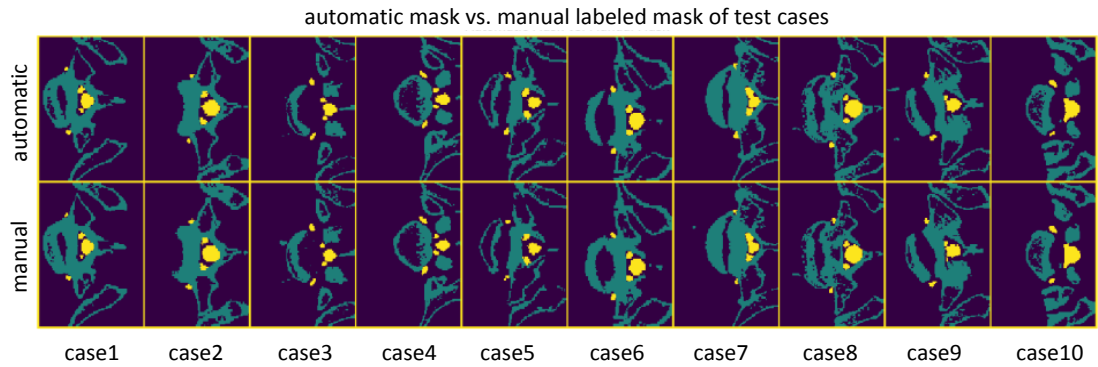

Figure 5 Automatic and mask manual labeled mask.

Table 2. Pixel accuracy of testing cases.

| Pixel Accuracy(%) | 1 | 2 | 3 | 4 | 5 | 6 | 7 | 8 | 9 | 10 | mean |
|---|---|---|---|---|---|---|---|---|---|---|---|
| bone | 89.7 | 85.8 | 82.0 | 99.5 | 99.9 | 89.5 | 99.8 | 98.5 | 96.9 | 98.9 | 94.0 |
| nerve | 94.2 | 85.3 | 89.1 | 92.7 | 85.8 | 92.9 | 94.4 | 93.9 | 94.2 | 91.8 | 91.8 |

Table 3. IoU of testing cases.

| IoU(%) | 1 | 2 | 3 | 4 | 5 | 6 | 7 | 8 | 9 | 10 | Mean |
|---|---|---|---|---|---|---|---|---|---|---|---|
| bone | 89.7 | 85.7 | 82.0 | 94.1 | 85.2 | 89.4 | 90.1 | 92.0 | 93.9 | 95.2 | 89.7 |
| nerve | 84.8 | 81.3 | 83.4 | 82.9 | 76.3 | 82.4 | 81.4 | 86.9 | 80.3 | 87.4 | 82.7 |

Table 4. Dice Score of testing cases.

| Dice Score(%) | 1 | 2 | 3 | 4 | 5 | 6 | 7 | 8 | 9 | 10 | mean |
|---|---|---|---|---|---|---|---|---|---|---|---|
| bone | 94.6 | 92.3 | 90.1 | 97.0 | 92.0 | 94.4 | 94.8 | 95.8 | 96.9 | 97.5 | 94.5 |
| nerve | 91.8 | 89.7 | 91.0 | 90.7 | 86.6 | 90.3 | 89.7 | 93.0 | 89.1 | 93.2 | 90.5 |

**Discussion and conclusion:**
Automatic segmentation of lumbosacral nerves may benefit quicker 3D reconstructions of lumbosacral structures, which will contribute to pre-procedure radiographic planning, spinal navigation and even robotic procedures in the near future. To the best of our knowledge, this is the first study automatically segmenting lumbosacral nerves on CT with deep learning.

Combined evaluation of lumbosacral structures (e.g. nerves, bone) on multimodal radiographic images is routinely conducted prior to spinal surgery and interventional procedures. Generally, MRI is conducted to differentiate nerves, while CT is used to observe bony structures. Unfortunately, a lot of information of soft tissue on CT is ignored. Nevertheless, deep learning may have the capacity to differentiate and segment all soft tissue structures on CT such as nerves, vessels, muscles, ligaments and so on. All these structures may play important roles in surgical planning, pre-interventional evaluation, and even in navigational and robotic surgery. MRI will probably work for segmenting soft tissue, and volumetric MRI with enhancement may be better for 3D reconstruction of soft tissue such as vessels and nerves. Many studies used diffusion tensor imaging (DTI) or MR neurography (MRN) technique to enhance the spinal nerves for diagnosis and surgical planning [25][26]. Yet, the scanning for DTI or MRN can be lengthy and costly. Also, the data size of volumetric MRI is inconveniently large, which is not applicable in practice. We have shown that thin-layer CT is a good candidate for segmenting nerves and reconstructing them into 3D even without any contrast. The current study disclosed a substantial potential of non-contrast CT in segmenting spinal nerves.

As more and more medical imaging datasets are created by medical experts, the application of deep learning in radiology is growing because of its excellent performance in recognition and segmentation. The performance of our model further supports the findings of other similar studies deploying U-Net and its variants for semantic segmentation of biomedical images [27][28]. Although small structures (e.g. nerve) tend to have lower dice score compared with large structure (bone) in semantic segmentation, our study still achieved a satisfactory Dice score of 0.905. Furthermore, SPINECT segments lumbosacral nerves and bones in about 3.1 seconds, which is much shorter than about 30 minutes for manual segmentation. In summary, the developed model has the potential to be adopted in work flow of spinal interventions and minimally invasive spine surgery.

The current study is not without limitations. Firstly, this pilot study only conducted segmentation on L5/S1 level, because it is the most difficult level for spinal intervention and minimally invasive spine surgery due to anatomic obstacles. SPINECT will be developed and tested on more levels (e.g. L3/L4, L4/L5) and different spinal region (e.g. thoracic, cervical) in the near future. Secondly, paraspinal structures such as vessels, muscles and even ligaments also play an important role in radiographic planning of spinal interventions and surgery, and even in navigational or robotic surgery. Semantic segmentation of multiple structures will be integrated into SPINECT. While the subject number and segmentation accuracy is acceptable, more cases may be needed and the accuracy shall be further improved.

In conclusion, deep learning with a 3D U-net can effectively segment spinal nerves and bones from CT. The results of this study suggest that our proposed SPINECT can be used to segment spinal nerves in CT seemingly within near human-expert performance.

**Contribution disclose:**
Guoxin Fan, William M. Wells and Shisheng He designed the study; Guoxin Fan and Yufeng Li conducted manual segmentation. Chaobo Feng and Dongdong Wang screened and extracted the data; Guoxin Fan and Huaqing Liu developed algorithm and wrote the manuscript. Huaqing Liu, Guoxin Fan and Zhenhua Wu validated and tested the model. William M. Wells III, Jie Luo, and Xiaofei Guan revised the manuscript. Shisheng He provided administrative support and approved the final version.

**Acknowledgement:**
This work is funding by CSC (201706260169) and NIH grant P41EB015898 (wmw). We would like to thanks the colleagues (Jagadeesan J, Mehrtash A, Zhou H, Kapur T, Kikinis R) in Surgical Planning Laboratory for professional comments and suggestion.